\pdfoutput=1 
\RequirePackage{fix-cm}
\documentclass{svjour3}                     
\smartqed  
\usepackage{graphicx}
\usepackage{amsmath}
\usepackage{color}
\usepackage[numbers]{natbib}
\usepackage{float}
\usepackage{caption}
 \journalname{arxiv}
\usepackage{xspace}
\usepackage{amssymb}
\usepackage{fontenc}
\newcommand\E{\ensuremath{\mathbb{E}}\xspace}
\newcommand{\acontrario}{\acontrario\xspace}
	{\ensuremath{\left \{ \hskip -1.5 mm \begin{array}{#1@{\quad}l}}}%
	{\end{array}\right.}

\newtheorem{mydefinition}{Definition}

\newcommand{\norme}[1]{\ensuremath{\Arrowvert #1 \Arrowvert}}

\newcommand\yy{\ensuremath{\mathbf{y}}\xspace}

\begin{document}

\title{Particle detection and tracking in fluorescence time-lapse imaging: a contrario approach
}


\author{Mariella Dimiccoli
 \and
        Jean-Pascal Jacob
\and Lionel Moisan}


\institute{Mariella Dimiccoli \at
              Computer Vision Center (CVC) and Universitat de Barcelona (UB)\\
              Tel.: +34 93 402 16 37\\
              \email{mariella.dimiccoli@cvc.uab.edu}           
           \and
           Jean-Pascal Jacob \at
              Paris Descartes University (Paris V)
	       Laboratory MAP5 (CNRS UMR 8145)
           \and
           Lionel Moisan \at
              Paris Descartes University (Paris V)
	       Laboratory MAP5 (CNRS UMR 8145)
}

\date{Received: date / Accepted: date}

\maketitle

\begin{abstract}
This paper proposes a probabilistic approach for the detection and the tracking of particles in fluorescent time-lapse imaging. In presence of very noised and poor quality data, particles and trajectories  can be characterized by ana-contrario model, that estimates the probability of observing the structures of interest in random data. This approach, first introduced in the modeling of human visual perception and then successfully applied in many image processing tasks,  leads to algorithms that do not require a previous learning stage, nor a tedious parameter tuning and are very robust to noise. Comparative evaluations against a well established baseline show that the proposed approach outperforms the state of the art.


\end{abstract}

\section{Introduction}
\label{intro}

\begin{figure}[h]
\centering
    \includegraphics[width=68mm]{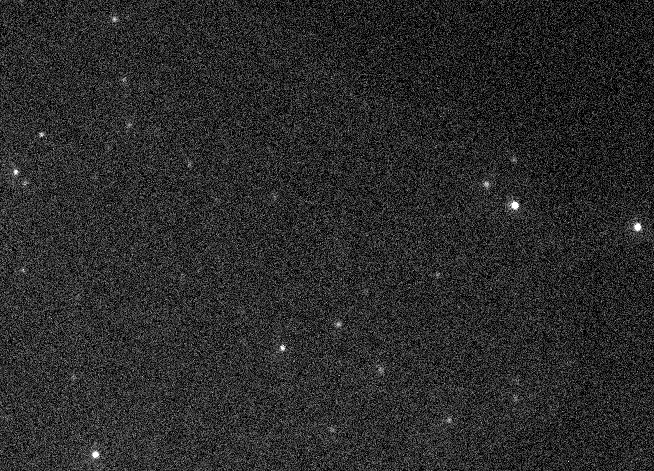}
    \caption{ 
    {\bf Example of particles corresponding to protein aggregates into bacteria cells visualized through a fluorescence microscopy.}
      }
  \label{fig:realSpots}
\end{figure}

Advances in microscopy and fluorescence technology over the last years, have led to the collection of huge amounts of fluorescent biological data, which require the use of automatic image processing tools to be analyzed quantitatively \cite{Akhmanova2008Tracking,Berginski2011HighResolution,Jandt2012Modeling,Agarwal2011ATPdependent,Coquel2013Localization}. 
However, mainly due to the poor data quality and the complexity of subcellular component dynamics, fluorescent time-lapse imaging represents a challenging domain for automatic image analysis. Indeed, when a single point-source of light is brought to a focus with a lens, the point-source image has a typical normalized intensity distribution called Point Spread Function (PSF), which can be very accurately approximated by a 2D Gaussian  \cite{thomann2002automatic}. In addition, in fluorescent imaging, the point-sources are often  subcellular structures, whose size is typically smaller than the resolution limit of the microscope, resulting in a diffraction limited spatial resolution. As a consequence, the objects of interest typically appear as bright spots severely blurred, commonly called \textit{particles} (see Fig.\ref{fig:realSpots}) over a possible widely varying background intensities.  Furthermore, specially in live cell imaging, the signal-to-noise ratio (SNR) is typically very low. Indeed, the intrinsic photon noise introduced in the imaging process can be reduced only by increasing the light intensity or the exposure time, which in turn causes the fading of the fluorescent signal, a process called photobleaching \cite{wu2010microscope}. To the poor data quality, the analysis of time-lapse sequences adds further challenges since it often requires to track over time multiple appearing/disappearing particles which undergo heterogeneous motion. In addition, tracking algorithms have to cope with missed detections and spurious particles that may arise from the detection step. 

At present, a large number of algorithms and software tools are available for the spatial detection and temporal linking of particle and, in recent years, a number of works have attempted to  address the problem of objectively assessing available methods under different experimental conditions \cite{Cheezum2001Quantitative,Carter2005Tracking,Smal2010Quantitative,Ruusuvuori2010Evaluation,Godinez2009Deterministic}, but they were limited to either one aspect of the task (particle detection or particle temporal linking) or to a single biological scenario. A recent work published on Nature Methods  by Chenouard et al. \cite{Chenouard2014Objective} has collected the results of an open competition  organized in 2012 to which participated $14$ teams. The challenge was to test tracking algorithms of each team on common simulated data sets, representative of different biological scenarios, and to evaluate their performances by using a common set of evaluation criteria. This study indicates that, at present, there exists no universally best method for particle detection and tracking since a method reported to work for certain experiments may not be the right choice for another application. In particular, it has been shown that most available tracking techniques cannot cope with high levels of noise and high particle density.

In this manuscript, we propose a probabilistic approach for the detection and the temporal linking of near-circular particles in biological images and we show its advantages over existing ones in terms of control of false alarms, robustness to noise, and reduced number of tuning parameters. In addition, being the approach unsupervised, it avoids the drawback of supervised methods such as the bias-variance dilemma and the need of a cumbersome learning stage. The contribution of this paper is twofold: first,  we propose a novel method for particle detection based on the a-contrario framework \cite{desolneux00}; second, using the baseline issue of the 2012 Particle Tracking Challenge, we evaluate the performances of a particle tracking method that takes the proposed method for particle detection as input and links particle in successive frames by a recently introduced method for particle temporal linking \cite{abergel14}, also based on the a-contrario framework.



In the next section, we introduce the state of the art on particle detection and  particle temporal linking. In section \ref{sec:acontrario} we recall the formalization of the a-contrario framework that will be used in section ~\ref{particledetection} and section ~\ref{particletracking} to explicit the a-contrario model for particle detection and particle tracking respectively. We devoted section ~\ref{exper} to the introduction and discussion of the experimental results. Finally, in section  ~\ref{conclusions}, we draw our conclusions.

\section{Related work}
\label{sec:relatedWork}

\subsection{Particle detection}

Particle detection methods can be broadly classified into supervised and unsupervised. 
Basically, supervised methods learn the particle model appearance from annotated training data consisting of positive and negative samples, whereas unsupervised methods assume some particle appearance model and rely on different filtering and detection techniques. Typically, in unsupervised methods, the particle model derives from Gaussian approximations of the PSF ~\cite{santos00,cheezum01,thomann02,thompson02,sage05} or from wavelet decompositions~\cite{olivo96,olivo02,zhang07}, or from feature-based approaches \cite{Smal2010Quantitative} or from mathematical morphology ~\cite{serra82,soille2013morphological,vincent93,smal08}.  

Methods deriving from a Gaussian approximation of the PSF include the Top-Hat Filter (TH) \cite{bright1987two,breen1991locating} and  the Spot-Enhancing Filter (SEF) \cite{Sage2005Automatic}.
TH \cite{bright1987two,breen1991locating} extracts small, compact or rounded objects from images. This is achieved by exploiting apriori information about object shape and predetermined information about their intensity from a circular interior region around a candidate point and a surrounding annular region. If the brightness difference in the two regions exceeds a threshold level, the candidate point is considered to be a particle. 
SEF was proposed by Sage et al. \cite{Sage2005Automatic} and basically consists of an enhancement filter resting on estimation theory, following which the maximum SNR detector of a given signal, or template, in additive stochastic noise is provided by the whitened matched filter. The matched filter is obtained by convolving the unknown signal with a conjugated time-reversed version of the template. The authors showed that the optimal detector of Gaussian-like particles in a fractal-like noise with a spectral power density that decays like $1/\omega^2$, where $\omega$ is the radial spatial frequency, is the Laplacian of a Gaussian, also known as Mexican hat filter. 

Methods deriving from wavelet decomposition include the Wavelet Multiscale Product (WMP) and the Multiscale Variance-Stabilizing Transform Detector (MSVTV). WMP is based on the wavelet decomposition  introduced in \cite{olivo02}, under which real objects, contrary to noise and randomly distributed data, are represented by a small number of wavelet coefficients that are correlated and propagated across scales. Hence,  objects can be detected simply by thresholding the multiscale product image. 
MSVTV is based on the multiscale variance-stabilizing transformation \cite{zhang07}.  Since objects under this transformation have to be localized in both space and frequency, large values of the transformed image usually correspond to some structure and smaller ones to noise.  The significant wavelet coefficients are detected by performing multiple hypothesis testing through the Benjamin-Hochberg procedure \cite{zhang07}. In the reconstructed image, only significant coefficients are nonzero and the background is largely removed whereas objects are preserved. Detection is performed by thresholding the reconstructed image. 

Methods deriving from morphological operators include the Grayscale Opening Top-Hat Filter (MTH) and H-Dome Based Detection (HD). The MTH \cite{serra82,soille2013morphological} is obtained by subtracting to the image its opened version, obtained by using a flat disk as  structuring element. The subtraction yields an image with only the removed objects which correspond to round light objects on a dark background. Contrary to the MTH, which select only compact structures smaller that the structural element, HD \cite{vincent93,smal08} acts  by subtracting from the original image $f$ the morphological reconstruction of the image $f-h$, where $h$ is a constant image.  This detector depends on the local contrast regardless the morphology or the scale of the objects.  A threshold above $h$ then only keeps the $h$-contrasted peaks (i.e local maxima). Its shortcoming is that small contiguous particles are extracted as one connected region because the size of the structuring element is wider than the minimum distance between the peaks of adjacent particles. Methods deriving from feature-based approaches are introduced and called Image Features Based Detection (IDF) in \cite{Smal2010Quantitative}. The key idea underlying these methods is to combine image intensities with local curvature information. This is achieved by computing at each pixel the determinant of the Hessian matrix with a smoothing scale \cite{romeny2003front} or, alternatively, by multiplying the value of the determinant of the Hessian with the intensity values. 
Supervised methods for particle detection include AdaBoost (AB) \cite{jiang2007detection}) and Feature Discriminant Analysis (FDA). Both approaches work by classifying image patches extracted from the images through a sliding window approach as particle or background. A set of four Haar-like features are extracted from each image patch.  The AB classifier consists of a sequence of weak classifiers which are combined in a weighted sum to create the final output of the boosted classifier.  FDA \cite{mclachlan2004discriminant} is a statistical technique that aims at finding, during training,  a projection where the class separation (particle and background) is maximized, taking into account the mean and the covariance matrix for each class.  This information is used during testing to generate a classification map, which convey particles when  thresholded at a value, which is also automatically estimated from training data.
From the outstanding quantitative comparison work made by Smal et al. \cite{Smal2010Quantitative}, which includes seven unsupervised (TH, SEF, WMP, MSVTV, MTH, HD and IDF) and two supervised methods (AB and FDA), differences in performances are negligible at high SNRs ($>5$) but performances of most methods drop out at SNR lower than 4. Taking into account also the number of parameters and the sensitivity of the methods to parameter changes, supervised methods achieve overall better performances at low SRNs but at the price of a cumbersome training stage, which may possibly introduce a bias.    

\subsection{Particle temporal linking}
\label{sec:relatedWorkTRacking}

Early methods for particle temporal linking have addressed the easier problem of tracking a single-particle but they may also track multiple particles whose trajectories are sufficiently separated in space \cite{Bohs1993Real,Bohs1993Real,Yildiz2003Myosin,Schutz1997Singlemolecule,Anderson1992Tracking,Kagawa2003Stepping}. Existing algorithms for tracking multiples particles can be broadly classified into deterministic and probabilistic approaches. Typically, deterministic approaches \cite{celler2013single,hager2004multiple,casuso2012characterization,rink2005rab,sbalzarini2005feature} act by first  estimating the position of each particle in each frame independently and then by linking particles in successive frames. Instead, probabilistic approaches \cite{chenouard2009multiple2,liang2010tracking,2001Sequential,coraluppi2011multi,Godinez2008Probabilistic,winter2012axonal} include a spatio-temporal filtering mechanism which allows to better exploit temporal information capturing the uncertainty of the measures due to noise (random variations) and other inaccuracies. 
Deterministic approaches may be local or global. Local approaches to link particles are  willing to fail when particles move quickly, close to each other and in presence of spurious/missed detections. Deterministic global strategies such as the global nearest neighbor approach \cite{Cox1993Review,casuso2012characterization,husain2012software} or Gaussian template matching  \cite{hager2004multiple} attempt to find and to propagate the single most likely hypothesis at each frame. In multiple hypothesis tracking \cite{reid1979algorithm}, temporal information is exploited to solve assignment ambiguities by delaying the association task between a set of measurements and a set of tracks to future observations that will resolve the conflict. This approach suffers from combinatorial explosion.
Typically, deterministic global strategies assume a motion mode of the particles and therefore are enable to deal with a variety of trajectory speeds at the same time. A deterministic global approach that does not assume motion modes was proposed by Sbalzarini and Koumoutsakos \cite{sbalzarini2005feature}: associations between points corresponding  to the same physical particle in subsequent frames are computed by minimizing a cost functional with topological constraints through a greedy algorithm. Another example of global deterministic approach was proposed by Jaqaman et al. \cite{jaqaman2008robust}. This method works by first linking particles between consecutive frames and then by linking the resulting track segments into complete trajectories. Both steps are formulated as global combinatorial optimization problems whose solution identifies the overall most likely set of particle trajectories throughout the sequence.



More sophisticated probabilistic multi-particle tracking algorithms model the object trajectory as a dynamic system, whose state-space evolution over time is  described through two equations: the measurement equation relating the observed data $Z$ to the state vector $X$  and the system transition equation for the state vector $X$. The state vector $X$ represents the object motion to be estimated and the measurement vector $Z$ represents the observed motion \cite{Arhel2006Quantitative,Isard1998CONDENSATION,2001Sequential}. A popular probabilistic approach is the Bayesian Sequential Estimation, which allows the recursive estimation of the so called filtering distribution $p(X_t|Z_{1:t})$,  describing the object state conditional on all the observations seen so far. If the posterior density at every time step is Gaussian and the system model along with the measurement model are linear, then the estimation can be done optimally by a Kalman filter \cite{Arhel2006Quantitative}. In contrast to Kalman Filter, Particle Filter \cite{Isard1998CONDENSATION,2001Sequential} exploits better the temporal information and allows to approximate models that are not linear and/or not Gaussian.  The filtering distribution is presented as a set of samples, or particles, with associated weights. The weights  are propagated over time to give approximations of the filtering distribution at subsequent time steps.  An important shortcoming of Particle Filters is that it leads to spurious detections. 
To address this problem, Godinez et al. \cite{godinez2011tracking} proposed to a probabilistic data association approach that combines a top-down strategy driven by the Kalman filter and a bottom-up strategy using standard localization algorithms for fluorescent particles. 
Other probabilistic approaches include probabilistic variations of multiple hypothesis tracking (MHT) \cite{willett2002pmht,coraluppi2011multi,chenouard2009multiple2,liang2010tracking}, multitemporal association tracking  \cite{winter2012axonal} and probabilistic data association \cite{shafique2005noniterative}. 
In probabilistic MHT \cite{willett2002pmht},  instead of assigning measurements to tracks as in traditional MHT algorithms, the probability that each measurement belongs to each track is estimated using a maximum a posteriori method. This algorithm has poor performances in cluttered environments. To solve this problem Chenouard et al. \cite{chenouard2009multiple2} proposed a MHT approach where detections are  linked to form target trajectories by using a Bayesian framework  aiming at building the set of tracks that maximizes the likelihood of the associations between tracks and measurements from the images of the sequence. In multitemporal association tracking \cite{winter2012axonal} the tracking problem is posed in a graph-theoretic framework where the detections are treated as vertices of a graph and the edges are possible inter-frame associations. Instead of finding associations between detected objects,  the association is found between feasible paths and the current set of tracks. The association is done by minimizing a cost function that approximates the Bayesian a posteriori probability estimate for the data association problem. The key difference with other approaches to solving the multitarget tracking problem is that it allows a single track to be assigned to more than one path. Shafique et al. \cite{shafique2005noniterative} perform probabilistc data association between set of particles belonging to different frames by looking for the maximum matching of a bipartite graph, where the partite set correspond to the set of points detected in two successive frames. The greedy algorithm has the advantage of allowing the use of different motion models and cost functions.  A major drawbacks of these methods are the assumptions about the probability distributions that do not necessarily hold and the large number of parameters.

A comparative evaluation of virus tracking methods has been done by Godinez et al. \cite{Godinez2009Deterministic}. They provided a performance evaluation of eight approaches suggesting that probabilistic approaches yield better performances than deterministic approaches.

\section{The a-contrario framework}
\label{sec:acontrario}
\begin{figure}[h]
  \begin{center}
    \includegraphics{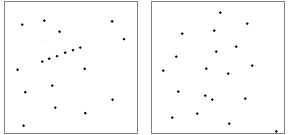}
    \caption{ 
    {\bf Illustration of Hemholtz principle.} 
  According to Helmholtz principle, we \textit{a priori} assume that the dots
    should have been drawn from an uniform distribution as in the right image. However, in left image, we perceive a group of aligned dots because such a structure is very unlikely to happen by chance in a random setting. Actually, also in the right image there is an alignment of three points but it does not pop out, because it is likely to happen by chance considering the total number of points.  
 }
  \label{fig:align}
  \end{center}
\end{figure}

The a-contrario framework rests on a perception principle stated by Helmholtz following which the human visual system detects structures in a group of objects when their configuration, according to one or several Gestalt laws, are very unlikely to happen by chance in a random setting. Basically, the a-contrario methodology requires two ingredients: a \textit{naive model}, that describes typical situations where no structure should be detected and one or several measurements defined on the structures of interest. 
For instance, when trying to discover alignments of points in an image, the \textit{naive model} should consist in a uniformly and independent draw of points where no alignments should be detected (see Fig. \ref{fig:align}). The measurements should define in what way an observation can be significant and are usually related to the visual saliency of the structure. For instance, for the image in Fig. \ref{fig:align}, the alignments of points should pop out because, assuming the \textit{naive model}, they are not likely to happen by chance considering the total number of points in the image. 
More formally, if the measurement function is high when the structure is pregnant, the amount of surprise when observing the measurement $x$ can be related to the probability $P(X > x)$, where $X$ is the random variable corresponding to the distribution of $x$ in the \textit{naive model}. We will usually have several measurements and in the classicala-contrario framework the amount of surprise will be measured by a Number of False Alarms (NFA), defined formally as follows.
\begin{mydefinition}[Number of False Alarms]
Let $( X_i )_{1 \leq i \leq N}$ be a set of random variables. A family of functions
$\big( F_i(x) \big)_i$ is a NFA (number of false alarms) for the random variables $(X_i)_i$ if
\begin{equation}
\forall \epsilon > 0, \quad \E\big( \# \{ i, \; F_i(X_i) \leq \epsilon \} \big) \leq \epsilon
\end{equation}
(as usual, the notation ``\#S'' stands for the cardinal of the set $S$).
\end{mydefinition}
The NFA ensures that the average number of detections made in the \textit{naive model} (false detections) at level  $\epsilon$ is less than $\epsilon$. Such detections are said meaningful.

\section{A-contrario particle detection}\label{particledetection}

\begin{figure}[h]
\centering
\includegraphics[height=3cm]{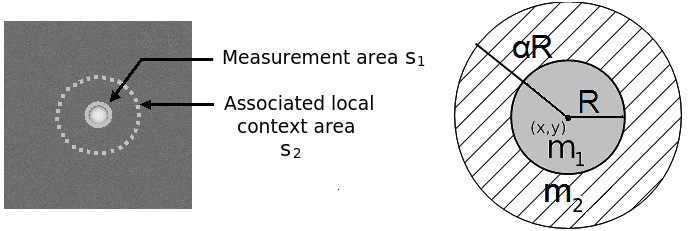}
\caption{Measurements taken at each test location: average intensity $m_1$ in the measurement area (gray disc centered at $(x,y)$), and average intensity $m_2$ in the associated local context area (a ring around the measurement area).}\label{patch}
\end{figure}

To apply the a-contrario framework to the detection of particles, we need to specify the \textit{naive model} $\mathcal{H}_0$ as well as a statistical measurement function $m$ able to characterize the visual saliency of the particles we are looking for. As \textit{naive model}, we take the realization of a Gaussian stochastic process with mean $\mu$ and standard deviation $\sigma$ ($\mathcal{H}_0 = \mathcal{N}(\mu; \sigma))$. In such random image, all image pixels are independent random variables with uniform probability distribution over some interval and therefore no structure should be detected. As measurement function to be performed at any given location $(x,y)$ of the image grid $\mathcal{T}$ , we consider the local contrast of a small patch centered at $(x,y)$ with respect to its local background. As observed by Grossjean and Moisan \cite{grosjean09}, modeling the local context is necessary to bypass the sensitivity of the observer to low frequencies in the detection task. To this goal, we define the statistical measure $m$ as the difference of one principal measurement, say $m_1$, defined on a disc of radious R centered at $(x,y)$ representing the detection area, and a context measurement, say $m_2$, defined on a ring surrounding the previous disc of radious $R$, with $(\alpha > 1)$ (see Fig. \ref{patch}).
Since a particle should be characterized by an high measurement function in the inner disc and a low measurement function in the outer disc, assuming that different particles are not too close so that the inner discs centered at them do not overlap, a particle detection occurs when the measurement $m1 - m2$ is high. In addition to cancel the low-frequency components of the single measure $m_1$, the measure $m$
yields detection thresholds independent of $\mu$, which is valuable when the precise value of $\mu$ is not known. In a sense, $m_2$ can be considered a local estimate of $\mu$.
The NFA that takes into account the local context is as follows.

\begin{mydefinition}[NFA for the model with contrast to the context]
Let $u$ be an image, let $\mathcal{T}$ be its grid and let $s_1$ and $s_2$ be two circular, concentric measurement kernels. A number of false alarms associated to the measurements $m_i(x,y) =  (u * s_i)(x,y) (i = 1, 2, (x, y) \in \mathcal{T}$ ) for the \textit{\textit{naive model}} $\mathcal{H}_0= \mathcal{N}(\mu,\sigma)$  is given by
\begin{equation}\label{eq:nfa}
NFA(\sigma,m_1,m_2) = |\mathcal{T}|\Phi_c \left( \frac{m_1-m_2}{\sigma||\phi_1 - \phi_2||}   \right)
\end{equation}

where $|\mathcal{T}|$ is the number of points of the image grid, $\mathcal{N}(\mu; \sigma)$ is the normalized two-dimensional Gaussian white noise, $\phi_1 = s_1* \mathcal{N}(\mu; \sigma)$ and $\phi_2 = s2*\mathcal{N}(\mu; \sigma)$ are the random variable following the \textit{naive model} distribution in $s_1$ and $s_2$ respectively that differ in the $\ell_2-$norm, and $\Phi_c(t) = \int_x^{+\infty} e^{-t^2} dt$ is the tail of the normal distribution.
\end{mydefinition}

We say that the $i-th$ point of the image grid $\mathcal{T}$ is $\epsilon-$meaningful, with $\epsilon> 0$, if $NFA(i) \leq \epsilon$.
In the \textit{naive model}, the random variable $M = \phi_1-\phi_2$ follows the law $\mathcal{N}(0,(1+\frac{1}{(\alpha^2-1)})\frac{\sigma^2}{\pi R^2})$. In practice, for a given $\epsilon$ that specifies the NFA and it is usually taken to be 1,  we perform at every pixel location $i$ the following test:
\begin{equation}
\frac{m_1-m_2}{\sigma}\geq T(\epsilon,R,\alpha).\label{eqSigmavar}
\end{equation}
where $\sigma$ is an estimation of the standard deviation of \textit{naive model} and $T(\epsilon,R,\alpha)= \frac{erfc^{-1}(\frac{2\varepsilon}{|\mathcal{T}|})}{\sqrt{\frac{\pi}{2}}\sqrt{1-\frac{1}{\alpha^2}}R}$.

In Equation \ref{eqSigmavar}, the standard deviation $\sigma$ is set once for all and generally has to be estimated. In presence of both high and low contrasted spots, this could be problematic since if the estimation of $\sigma$ is bigger than the real noise standard deviation, low contrasted particles would be missed. For the opposite case, since high intensity values have a larger dynamic range than low intensity values, as soon as the measure $m$ becomes positive, it is very likely that $m$ becomes big compared to the global estimate of $\sigma$. In these conditions, even in locations very bad centered near a real particle, the NFA is typically small leading to very widespread detections, while the only meaningful detection is the center of the particle. To solve this problem and to allow the detection, without any bias, of low and high contrasted particles, we relaxed the weight due to the contrast by replacing the global estimation of $\sigma$ in Eq. \ref{eq:nfa}  by a local estimation of $\sigma$ in a local neighborhood on the candidate particle, say $\sigma_l$. This choice leads to a redefinition of the \textit{naive model}, that becomes a Gaussian white noise with a local standard deviation $\sigma_l$. 

\begin{figure}[h]
\begin{tabular}{ccrr}
\includegraphics[scale=0.5]{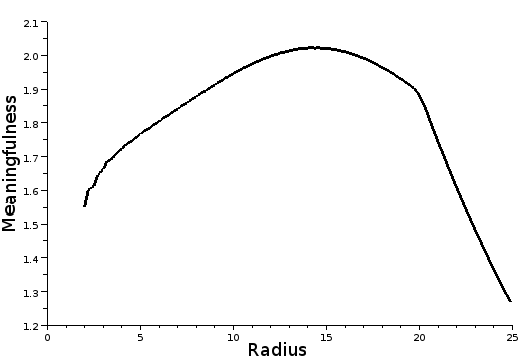}&
\includegraphics[scale=0.6]{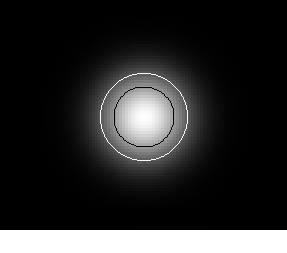}
\\ (a) & (b)
\end{tabular}
\caption{Left: measures of meaningfulness on a perfect Gaussian particle of $\sigma=10$ pixels. Right: corresponding maximum of the gradient (in black) and of the function $\frac{m_1-m_2}{\sigma_l}(R)$ (in white, $R_{opt}=1.45\sigma$)}\label{courbesgauss}
\label{fig:courbeGauss}
\end{figure}


\subsection{Particle spreading estimate and hiding process}

The proposed detection approach leads to a natural particle spreading estimation which allows to quantify the amount of fluorescence of a particle. For a given particle location, once $\alpha$ is set,  the ratio $\frac{m_1-m_2}{\sigma_l}$ is a function of $R$,  the inner radius of the model. This function depends solely on the local contrast and its maximum gives an estimate of the particle spreading. In fact, as it can be observed in Fig.  \ref{fig:courbeGauss},  for a $2D$-circular Gaussian particle, the maximum of the ratio $\frac{m_1-m_2}{\sigma_l}$ is proportional to the spread of the Gaussian. Let 
\begin{equation}
R_{opt}=\underset{R}{argmax}(\frac{m_1-m_2}{\sigma_l}(R))
\end{equation}

be the estimation of the particle spreading. 

\begin{figure}[h]
\centering
\includegraphics[height=3.5cm]{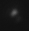}
\includegraphics[height=3.5cm]{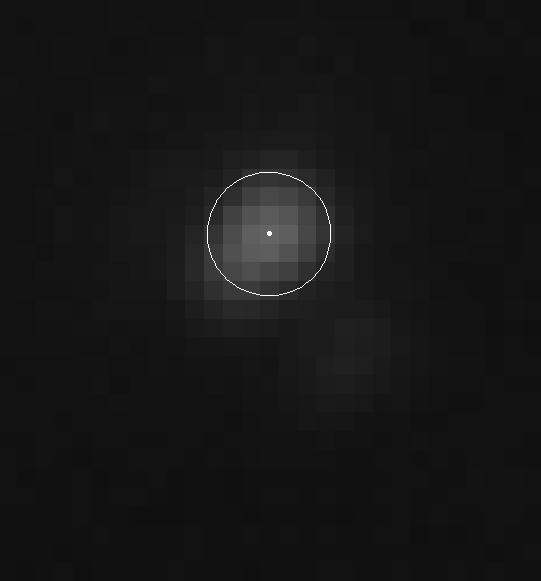}
\includegraphics[height=3.5cm]{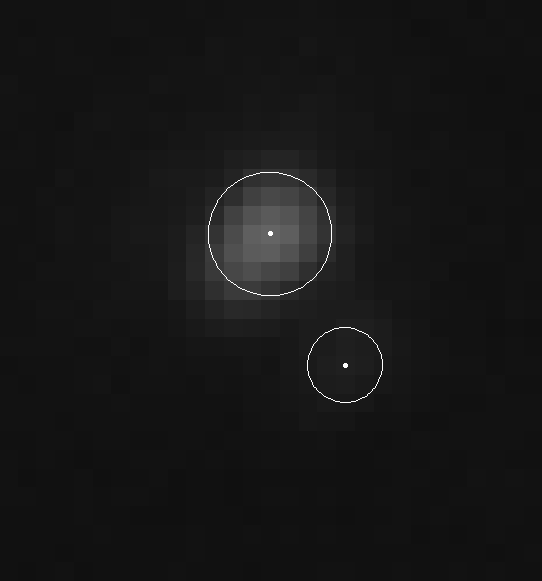}
\caption{(a) Original image. (b) $0.1$-meaningful detection. (c) 2-pass $0.1$-meaningful detections with hiding process ($R=2.5$ pixels, $\alpha=2$)}\label{seg85ex}
\end{figure}

The function $\frac{m_1-m_2}{\sigma_l}(R)$  is drawn in Fig.\ref{courbesgauss} (a)  for a local model centered on a 2D-circular Gaussian function with standard deviation $\sigma = 10$. In this case $R_{opt} = 1.45$. On real images, we compute the particle center locations for a given model and a given $\epsilon$ as the gravity center of $\epsilon$-meaningful connected components. The function $\frac{m_1-m_2}{\sigma_l}(R)$ is computed on each detected center and its maximum $R_{opt}$ gives a radius estimation.

The particle spreading estimation is also useful to improve the algorithm performances: the detection of low contrasted particles may fail when they are very close to large and high contrasted particles. To overcome such a problem, we consider that almost all  fluorescence values of a given particle $(x,y)$ are contained in the $95\%$ confidence interval, i.e inside a circle with a radius $R_{2\varsigma} = 2/1.45R_{opt}$. Then the algorithm removes any detected particle inside the $95\%$ confidence interval and redoes the detection (see Fig. \ref{seg85ex}). Practically, pixel values inside the circle $\textbf{C}(x, y,R_{2\varsigma} )$, when computing $m_1$, $m_2$ and $\sigma_l$ on close candidate particle locations, are set to the value of the background. Since this process deforms the pattern of the local model near former detections, the minimal distance between a hidden particle and the inner part of a new local model should be at least one pixel.

\begin{figure}[h]
\centering
\includegraphics[width=\textwidth]{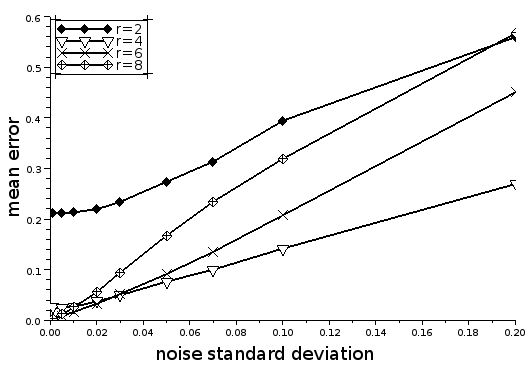}
\caption{Mean localization over 300 synthetic images as a function of the standard deviation of the Gaussian white noise.}\label{subresolution}
\end{figure}
\subsection{Sub-pixel refinement}
In fluorescence imaging the target of interest is usually smaller than the pixel size and all that is observed is the instrument's PSF centered on the particle's position that typically spreads several pixels wide. Therefore, in the study of particle temporal dynamics, it is of crucial interest to provide efficient solutions for sub-pixel detection. For instance, in the study of bacteria aging, the molecular components of interest are protein aggregates accumulated near bacteria boundaries. This particular location makes very ambiguous the correspondence between aggregates and cells, and sub-pixel accuracy becomes crucial to disambiguate this association. To achieve subpixel accuracy, we propose to refine the location of each detected particle, say $x$, by computing a weighted average of pixel coordinates in a circular neighborhood. The sub-pixel location, $\hat{x}$ is given by the following equation 

\begin{equation}
\hat{x} = \frac{1}{|M_x|}\sum_{y \in M_x} u(y)w(y)
\end{equation}

where $M_x$ is the circular neighborhood of the particle $x$ having cardinality $|M_x|$, $w(y) = u(y) - m_x$ if $u(y) - m_x > 0$, $w(y) = 0$ otherwise, being $m_x$ is the median intensity value of all pixels belonging to $M_x$. The median value $m_x$ is an estimation of the local background intensity: only pixels having intensities bigger than the local background intensity are considered in the weighted average. This avoids to include into the average pixel values corresponding to very close particles. To validate the proposed sub-pixel refinement, we considered a set of 300 synthetic images with a single particle at a sub-pixel location $x$, simulated by a Gaussian bi-dimensional signal of standard deviation $\sigma_s$ and noised with a white noise of standard deviation $\sigma_n$. We computed the mean error over the set of images as a function of the parameter $\sigma_s$ and of the radius $r$ of the circular neighborhood $M_x$. Using a radius $r = 6$, we achieved a precision of $1/10$ of pixel on synthetic images even on very poor contrasted particles (see Fig. \ref{subresolution}).

\section{A-contrario particle temporal linking}\label{particletracking}
the a-contrario model for particle temporal linking, first introduced in \cite{primet12} as ASTRE, estimates the probability of observing a trajectory in random data through a probabilistic criterion which combines data statistics, such as the number of images and particles, with trajectory  characteristics such as trajectory length, particle density, and smoothness. This criterion is used to drive a dynamic programming algorithm \cite{Bellman1952Theory}  which sequentially extract the most meaningful trajectories globally in time, while guarantying that no trajectory will be found in random data.
In the following we give a brief introduction to this method and to its more recent variant, called CUTASTRE \cite{abergel14}.

\subsection{Expliciting the a-contrario model for trajectory detection}

As for any othera-contrario-based method, the a-contrario approach for the detection of trajectories, is grounded on two elements: the \textit{\textit{naive model}} and  a statistical measurement function able to characterize the visual saliency of the trajectories. 
In the following, we assume that we are given $K$ images $I_1, ..., I_K$, each image $I_k$ containing $N$ points $X_1^k , ...,X^k_N$ corresponding to particles that have already been detected in each image of the sequence. We assume that the detections are noised so that the observed points may correspond to spurious particles and that some particles could have been missed.
Intuitively, by Helmholtz principle, we should not see trajectories appearing in the realizations of the \textit{\textit{naive model}}. 
In the case of trajectory detection, the \textit{naive model} is an uniform and iid draw of $N$ points in each of the $K$ frames.
Before introducing the statistical measurement function, we define the structures of interest.
\begin{mydefinition}[Trajectories without holes]
A \textit{trajectory} $T$ of length $\ell$ starting at frame $k_0$ is a tuple
$T = (k_0, i_1, ..., i_\ell)$, where $1 \leq i_p \leq N$ for all $p$ and $1 \leq \ell \leq K - k_0 + 1$.  We will denote by $\mathbb{T}$ the set of all trajectories.
There is a natural equivalence between a trajectory $T \in \mathbb{T}$ and the tuple of variables $X_T = (X^{k_0}_{i_1}, ..., X^{k_0 + \ell - 1}_{i_\ell})$ that we shall therefore sometimes abusively call a (random) trajectory too.
\label{traj1.def}
\end{mydefinition}

The statistical measurement function associated to a trajectory $T = (X^{k_0}_{i_1}, X^{k_0 +1}_{i_2}, X^{k_0 + l -1}_{i_l})$ with length $l$, where $X^k_i$ is the $i$-th point of frame $k$, is its maximal acceleration
$a_{\mathrm{max}}(t) = \max_{3 \leq i \leq \ell-1} \norme{\yy_{i+1} - 2
\yy_i + \yy_{i-1}}$.
The amount of surprise when observing a trajectory $T$ of length $l$ and acceleration $\delta:= a(t)$ is estimated by the upper bound 
\begin{equation}
\mathbb{P}_{\mathcal{H}_0}(a(T_l)\leq \delta) \leq (\pi \delta^2 /| \Omega|)^{l-2}
\end{equation}
where $\Omega$ is the image domain. The NFA can than be computed thanks to Lemma 1 in \cite{grosjean09}.

Trajectories are extracted iteratively. At each iteration, the value of the trajectory with minimal NFA among all trajectories, say $m$, is computed through a dynamic programming strategy and compared to  $ \epsilon$, the maximal NFA value of a trajectory that the user wants to extract (usually one chooses $\epsilon = 1$). If $m \leq \epsilon$, all points corresponding to the trajectory are removed from the sequence. This process is repeated until no trajectory with NFA less than $\epsilon$  can be found anymore. The ASTRE algorithm is global in time and has an unique parameter $\epsilon$. However, it has quadratic time and memory complexity  with respect to the number of frames, that may be prohibitive for some applications  involving long image sequences (say more than 1000 frames). Abergel and Moisan \cite{abergel14} proposed a modified version of ASTRE, called CUTASTRE with $\mathcal{O}(K)$ complexity, which preserves the rigorous control of false detections in pure noise offered by ASTRE achieving very similar performances. This important complexity reduction is obtained at the cost of two additional parameters: the temporal chunk and overlap sizes, that  in general, as proved in \cite{abergel14}, are easy to be set since their are related to the smoothness of the trajectory. In addition, CUTASTRE is not currently able to deal with trajectories with holes as ASTRE does. 

\section{Experimental Results}
\label{exper}
\subsection{Comparative evaluations of particle detection}
To evaluate quantitatively the performances of the proposeda-contrario particle detection method, we used the same experimental setup proposed by Smal et al. \cite{Smal2010Quantitative}. 
In this work, seven unsupervised (WMP, MSVST, TH, MTH, SEF, HD, IDF) and two supervised methods (AB and FDA)  were compared in terms of true positive and false positive rate, taking into account also the methods's sensitivity to parameter changes and data quality.

\begin{table}[H]
\caption{Optimal NFA parameter and corresponding performances of the a-contrario particle detection for different SNR} 
\vspace{3mm}
\centering 
\begin{tabular}{c c c c} 
\hline 
Image Type & NFA & TPR  & FPR\\ [0.5ex] 
\hline
& SNR = 4  \\
\hline 
A & 0.3 & 1 & 0.005  \\ 
B & 0.3 & 0.96 & 0.036  \\
C & 0.1 & 1 & 0.005 
\\
\hline 
& SNR = 3\\ [0.5ex] 
\hline 
A & 0.3 & 1 & 0.009  \\ 
B & 0.3 & 0.98 & 0.031  \\
C & 0.1 & 1 & 0.006  \\
\hline 
& SNR = 2\\
\hline
A & 0.3 & 1 & 0.008  \\ 
B & 0.3 & 0.99 & 0.028  \\
C & 0.1 & 0.97 & 0.005  \\
\end{tabular}
\label{table:acontrarioMultipleSNR} 
\end{table}

\begin{table}[H]
\caption{Radii parameters used for each type of image for all SNR} 
\vspace{3mm}
\centering 
\begin{tabular}{c c c } 
\hline 
Image Type &  R & $\alpha$ \\ [0.5ex] 
\hline 
A & 3 & 2   \\ 
B & 2 & 3.5   \\
C & 3 & 1.25  \\
\hline 
\end{tabular}
\label{table:radiiParameters} 
\end{table}

\subsubsection{Simulated image dataset and performance measures}
We used the publicly available ImageJ plugin \textit{syndata.jar} to generate three types of image sets,  each for SNR ranging from 2 to 4. Each type of image set corresponds to an uniform background (type A), a gradient background (type B) and a non-uniform background (type C). The performances of the a-contrario detection methods were evaluated by accumulating the numbers of true positive (TP)  and false negative (FN) for 16 images, each containing 256 ground truth objects and averaging the results over all objects. As in \cite{Smal2010Quantitative}, the tolerance was fixed to 4 pixels. To compare the algorithms were used two main measures: 1) the true-positive rate ($TPR$) defined as $TPR = NTP/(NTP + NFN)$, where $NTP$ stands for number of true positives and $NFN$ is the number of false negatives defined as $NFN=N0-NTP$, being $N0$ the number of objects in the ground truth; 2) the modified false positive rate defined as $FPR^* = NFP/(NTP+NFN)$, where $NFP$ is the number of false positives.  The value of TPR for the optimal parameters is denoted as $TPR^*$. In addition to detection performances with optimal parameters, the sensitivity to parameter changes and data quality is also considered. The sensitivity to parameter changes is evaluated by plotting the FROC curves obtained at a fixed SNR for two different values of a given parameter and for each image type. The sensitivity to data quality is quantified by considering the variation of TPR and FPR for different values of the SNR, keeping fixed the algorithm's  parameters for each image type.

\subsubsection{Discussion}
In Table \ref{table:acontrarioMultipleSNR} are shown the performances of the proposed method for the optimal NFA parameter for varying SNRs. These results have been obtained using fixed values of the internal radius $R$ and $\alpha$ for each image type, which are reported in Table \ref{table:radiiParameters}. These experiments demonstrate that the proposed method is robust to variations of data quality.

\begin{figure}[H]
\centering
\begin{tabular}{crrrrrr}
\includegraphics[height=7.5cm]{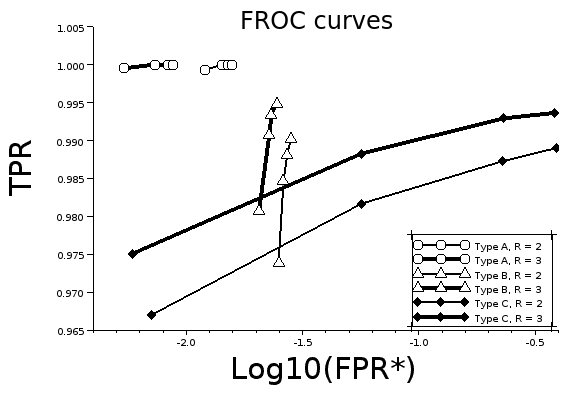}\\
(a)\\
\includegraphics[height=5.5cm]{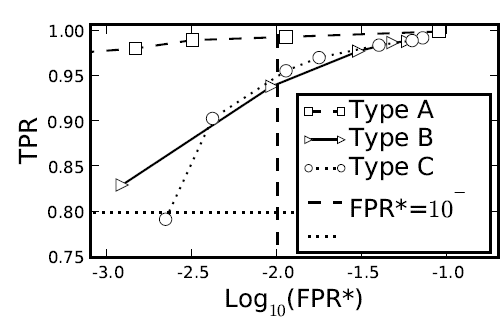}\\
 (b)\\
\includegraphics[height=5.2cm]{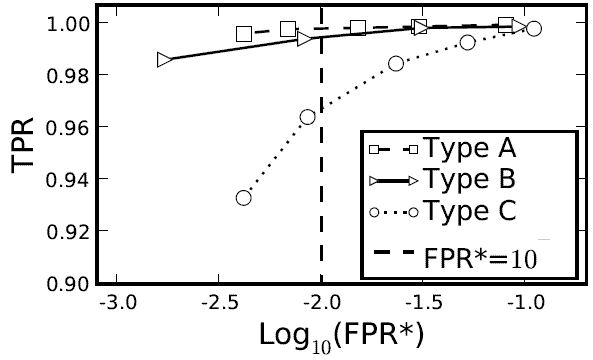}\\
 (c)
\end{tabular}
\caption{(a) FROC curves for the a-contrario method depending on the value of the R and $\alpha R$ at SNR = 2 and optimal value of the NFA. (b) FROC curves for AB method depending on the value of the $v_d$ threshold at SNR = 2 and with number of Haar-like features $N_{AB}=50$. (c) FROC curves for FDA method depending on the value of the $v_d$ threshold at SNR = 2. Image source for (b) and (c) \cite{Smal2010Quantitative}.} . \label{seg85ex}
\label{fig:RFOC_R_alpha}
\end{figure}

To evaluate the sensitivity of the proposed method to parameter changes, in Fig.\ref{fig:RFOC_R_alpha} (a) we report the FROC curves obtained at $SNR =2$ for two different values of the internal radius $R$ for each image type, varying $\alpha$. 
As it can be observed, the performances are quite stable with respect to variations of these parameters and performances do not go below the $96\%$ TPR.
In Table \ref{table:comparisons}, we compare our results to those obtained by using FDA and ADABOOST that, in the comparison work of Smal et al. \cite{Smal2010Quantitative},  reported the highest TPR* and the lowest sensitivity to parameter changes and data quality on our same dataset with respect to seven unsupervised methods (TH, SEF, WMP, MSVTV, MTH, HD, IDF) . 
The sensitivity of the measures $TPR$ and $FPR^*$ to a parameter, say $l_d$ (that correspond to the NFA for our method and to the threshold on the size of the clusters for FDA and ADABOOST) is measured through the values $S_T = -\partial{TPR}/\partial{l_d}$ and  $S_F = -(\partial FPR^*/\partial l_d)$ at the value of the parameter for which the $FPR^* = 0.01$ (only $1\%$ false positives) hereafter called \textit{optimal parameter}. As it can be observed, the sensitivity of our method to variations of the NFA is very small and order of magnitude smaller than the one of AB and FDA.

The proposed method has been extensively tested on biological images to detect protein aggregates in growing bacteria cultures as reported in \cite{Coquel2013Localization}. In this work,  to  deal with the presence of particles having variable size, we used a multiscale approach that in turn consists in using two different sizes for the inner radius.  In such experiments, the NFA was fixed to $1$, $\alpha=2$ and inner radii $R$ varying from $2$ to $3$ pixels. The same process was performed on four different growing sequences of about $90$ images comprising $1$ or $2$ particles at the beginning and up to about $150$ particles at the end. In mean $15$ particles per sequence were recovered thanks to the hiding process.




\begin{table}[H]
\centering
\begin{tabular}{c c c c c} 
\hline 
& ACONTRARIO\\
\hline
Image type & \textit{NFA} & TPR & \textit{$S_T$} & \textit{$S_F$} \\
\hline
A & 0.3 & 1 & $10^{-5}$ & $10^{-3}$ \\
B & 0.3 & 0.99 & $10^{-5}$ & $10^{-2}$\\
C & 0.1 & 0.97 & $10^{-5}$ & $10^{-2}$ \\
\hline 
& ADABOOST\\
\hline
Image type & \textit{$v_d^*$} & TPR & \textit{$S_T$} & \textit{$S_F$} \\
\hline
A & 3 & 0.99 & $10^{-3}$ & $10^{-3}$ \\
B & 31 & 0.94 & .01 & $10^{-3}$ \\
C & 30 & 0.94 & .01 & $10^{-3}$ \\
\hline 
& FDA\\
\hline
Image type & \textit{$v_d^*$} & TPR & \textit{$S_T$} & \textit{$S_F$} \\
\hline
A & 4.6 & 0.99 & $10^{-5}$ & .01 \\
B & 8.8 & 0.99 & $10^{-3}$ & .01 \\
C & 9.8 & 0.96 & $10^{-2}$ & .01 \\
\hline
\end{tabular}
\caption{Optimal parameters for each image type at SNR = 2. $v_d^*$ is the threshold on the size of the clusters used by ADABOOST  and FDA; NFA is the threshold used by the proposed method.}\label{table:comparisons}
\end{table}

From the above experiments, we can conclude that the proposed approach overall outperforms both supervised and unsupervised methods considered in \cite{Smal2010Quantitative} in terms of true positive ratio with only $1\%$ false positives at very low SNR and in terms of robustness with respect to data quality and parameter changes. In addition, the proposed method has the advantage of not requiring a cumbersome training stage on similar data, which requires additional parameters to be estimated such as the number of Haar-like features.  Finally, it has the advantage of allowing the user to directly set the parameter $\epsilon$, which represent a bound of the average number of detections that would be made in pure noise data (that is on spurious detections). Due to its robustness to parameter changes and to poor data quality, as well as to its unsupervised nature, our method is very suited to be used by biologists. In the next section we further validate the suitability of the proposed method as input for a particle temporal linking algorithm.

\subsection{Comparative evaluations of particle detection followed by temporal linking}

In this section, we evaluate the performances of the a-contrario temporal linking approach called CUTASTRE introduced in  \cite{abergel14} when particles are detected by using the a-contrario approach proposed in this paper. Our goal is twofold: first, to validate the suitability of the proposed particle detection method as input for the task of linking particles in successive frames; second,  to demonstrate the advantages of the a-contrario appraoch for both particle detection and temporal linking. For that, we used the baseline issue of the 2012 Particle Tracking Challenge data (see http://www.bioimageanalysis.org/track) to which participated 14 teams. Each team applied his method independently on a common dataset and evaluated the results using a set of commonly evaluation criteria.

\subsubsection{Simulated image data sets}

Since the ground truth is generally not available for real biological data and manual annotation by human observers is subjective, costly and prone to bias \cite{huth2010significantly}, the authors simulated image data for the challenge. The simulated data, together with their corresponding ground truth, take into account three factors that usually have large influence on tracking results: the particle dynamics characterizing a biological scenario, the particle density in a fixed field of view and the particle signal relative to noise. Four biological scenarios were simulated, including near-circular particles showing a Brownian motion (VESICLES), near-circular particles switching between Brownian and directed motion models (RECEPTORS), near-circular particle captured in 3D switching between Brownian and directed motion models (VIRUS), elongated particles showing near-constant velocity motion (MICROTUBLES). For each biological scenario three levels of particle density were considered (low $~$100 particles, mid $~$500 particles, high $~$1000 particles), and four SNRs (1,2,4,7). Being the proposed algorithms suitable only for 2D detection and tracking of symmetric Gaussians, we considered only 2 of the 4 aforementioned scenarios, VESICLE and RECEPTOR,  amounting to a data set consisting of 24 image sequences. It is worth to mention that the 3D scenario represented by VIRUS is the easiest from a temporal-linking  point of view since these particles shows a near-constant velocity motion and VIRUS show the same motion mode than RECEPTORS, being the only difference the shape of the particles.

\subsubsection{Performance measures}
A set of five complementary criteria that give a complete and intuitive characterization of the tracking results when tracking with a varying number of particles in a cluttered environment were used to evaluate the tracking performances. In the following, we give an intuitive explanation of each of them and we refer the reader to the work of \cite{Chenouard2014Objective} for further details.  Let us define a track $\theta$ that exists from time $t_{start}$ to time $t_{end}$ as a temporal series of subsequent spatial positions, say the set $\theta= \{ \theta(t) = (x(t),y(t))\}$, with $t= t_{start},...,t_{end}$.

The \textbf{distance between two tracks} is defined as 
\[
\alpha(\theta_1,\theta_2) = \sum_{t=0}^{T-1} ||\theta_1(t) - \theta_2(t)||_{2,\epsilon}
\]
where $T$ is the lenght of the image sequence and the distance between two positions is defined as 

\[
||\theta_1(t) - \theta_2(t)||_{2,\epsilon} = \min(||\theta_1(t) - \theta_2(t)||_{2},\epsilon)
\]
where $|| \cdot ||$ is the standard $\ell_2$ norm of $\mathcal{R}^2$ and $\epsilon \in \mathcal{R}_+$.
This measure limits the penalization for tracks that are more than $\epsilon$ apart to $\epsilon$. The parameter $\epsilon$ was fixed to five in the particle tracking challenge.

The \textbf{distance between two track sets} is defined as follows. Let $X$ be the set of ground-truth tracks and, $Y$ the estimated set of tracks and $\tilde{Y}$ the extended version of $Y$ with dummy tracks. Denoting by $\Omega$ the set of tracks that can be obtained by taking $|X|$ elements from $\tilde{Y}$, the distance between $X$ and an element $Z$ from $\Omega$ is defined as the sum of the distances between $|X|$ pairs given by the ordering of the two sets. The distance between $X$ and $Y$ is then defined as the minimum distance between $X$ and all possible $Z$:
\[
d(X,Y)= \min_{Z\in\Omega} \sum_{k=1}^{|X|} d(\theta_k^X,\theta_k^Z)
\]
By using the above definitions of distance, five metrics used for evaluation are defined as follows: 

\begin{enumerate}
\item The normalized score: $\alpha(\theta_1,\theta_2) = 1 - d(\theta_1,\theta_2)/d(\theta_1,\Phi)$
where $\Phi$ denotes the set of $|\theta_1|$ dummy trucks. It measures the overall degree of matching of groundtruth and estimated tracks without taking into account spurious tracks. 
\item The criterion $\beta(\theta_1,\theta_2) = \frac{d(\theta_1,\Phi)+d(\theta_1,\theta_2)}{d(\theta_1,\Phi)+d(\bar{\theta_2},\Phi)}$
 that measures the overall degree of matching of groundtruth and estimated tracks with a penalization of nonpaired estimated tracks.
\item Jaccardi similarity index for positions $JSC = \frac{TP}{TP+FN+FP}$. It lies in the interval $[0,1]$ and takes value 1 only if there are not spurious tracks in $Y$ and all positions pairs in $(X,Z^*)$ are matching. 
\item Jaccard similarity coefficient for entire tracks
$JSC_{\theta} = \frac{TP_{\theta}}{TP_{\theta}+FN_{\theta}+FP_{\theta}}$. It lies in the interval $[0,1]$ and takes value 1 only if there are not spurious tracks in $Y$ and $Z^*$ does not contain dummy tracks.
\item Root Mean Squared Error (RMSE) that indicates the overall localization accuracy of matching points in the optimally paired tracks by using the Euclidean distance
\end{enumerate}

\subsubsection{Discussion}
In this section we discuss comparative performances of the a-contrario approach against 14 methods that partecipated to 2012 Particle Tracking Challenge. 
Since in the dataset only trajectories without holes are considered, we used CUTASTRE \cite{abergel14} which cannot handle trajectories with holes but has linear complexity. For particle detection we used a multiscale approach consisting of using different values of the internal radii (from 1.5 to 3). We varied the algorithm parameters in a small range and kept the best result for each scenario. 

\begin{figure}[H]
\begin{tabular}{ccrrrrrrrrrrrrrrr}
\includegraphics[width=8cm]{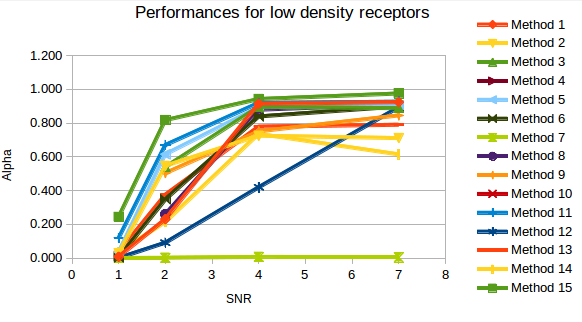}
&\includegraphics[width=8cm]{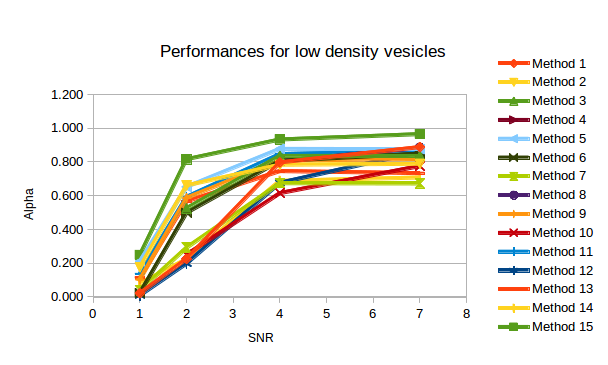}
\\
\includegraphics[width=8cm]{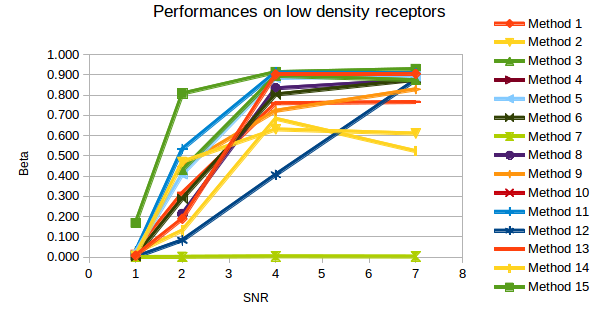}
&\includegraphics[width=8cm]{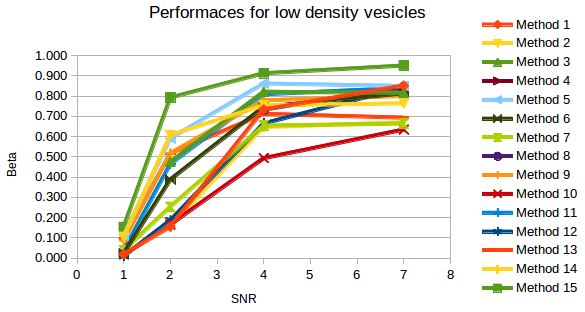}
\\
\includegraphics[width=8cm]{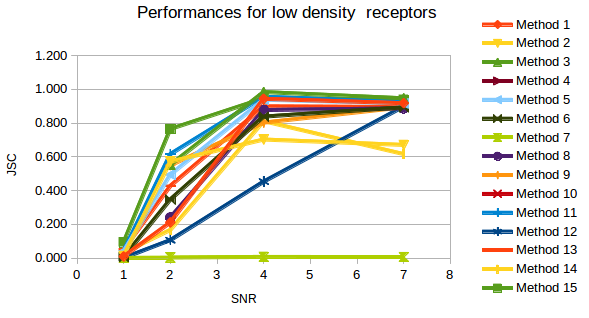}
&\includegraphics[width=8cm]{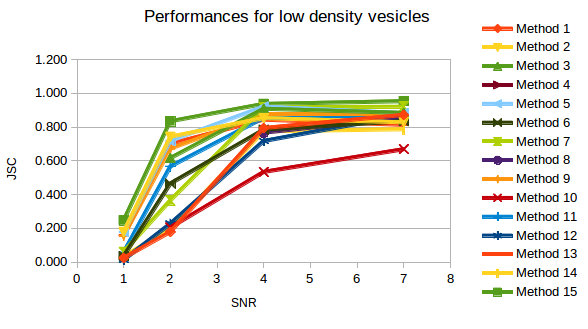}
\\
\includegraphics[width=8cm]{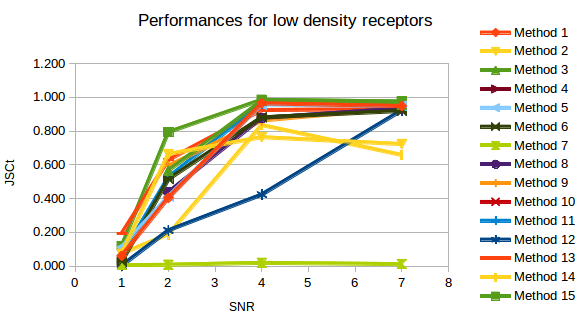}
&\includegraphics[width=8cm]{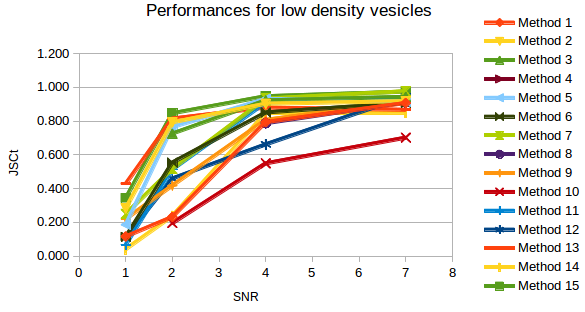}
\\
\includegraphics[width=8cm]{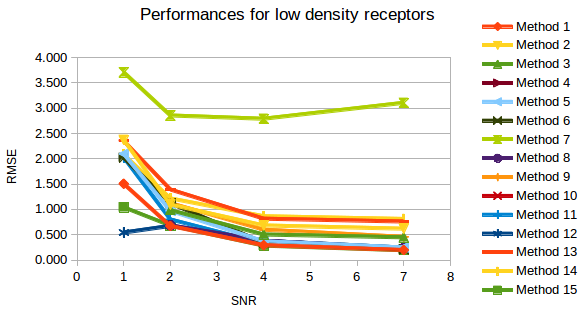}
&\includegraphics[width=8cm]{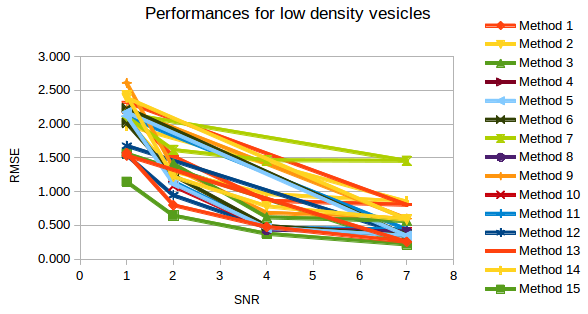}
\\ (a) & (b)
\end{tabular}
\caption{Performance measures for low-density particle scenarios: (a) RECEPTORS, (b) VESICLES. From up to down are show the metrics $\alpha$, $\beta$, $JSC$, $JSC_t$ and $RMSE$. Method 15 is the proposed approach.}
\label{lowDensity}
\end{figure}

\begin{figure}[H]
\begin{tabular}{ccrrrrrrrrrrrrrrr}
\includegraphics[width=8cm]{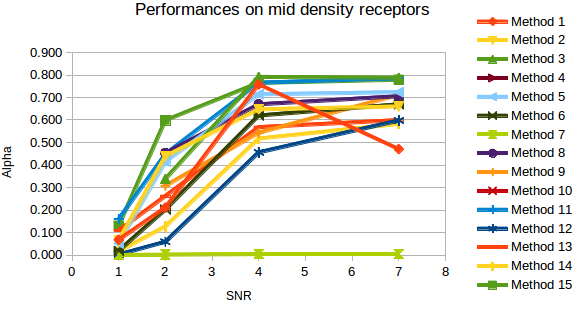}
&\includegraphics[width=8cm]{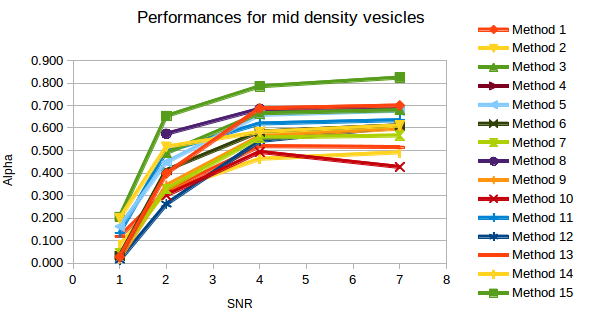}
\\
\includegraphics[width=8cm]{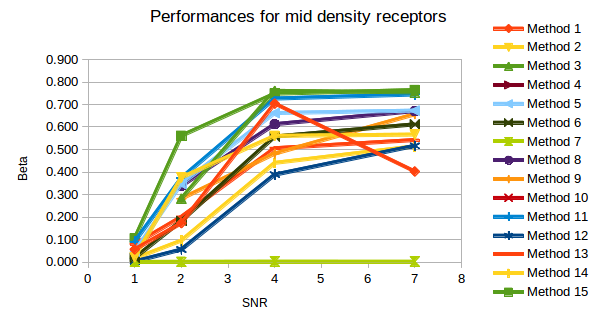}
&\includegraphics[width=8cm]{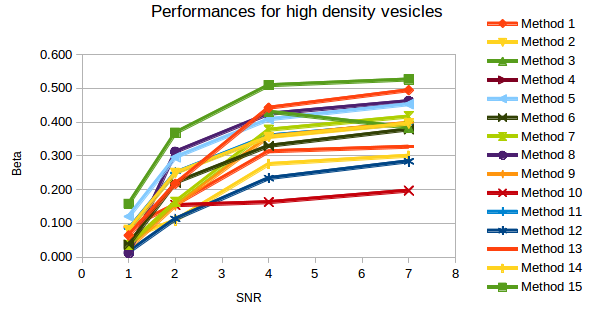}
\\
\includegraphics[width=8cm]{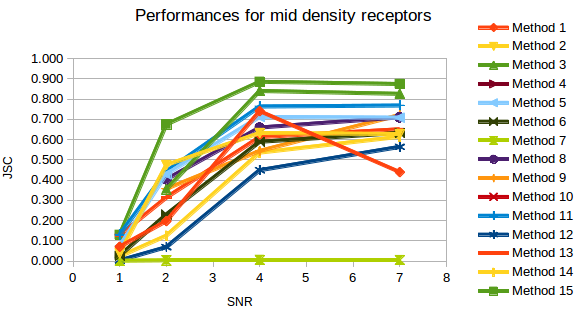}
&\includegraphics[width=8cm]{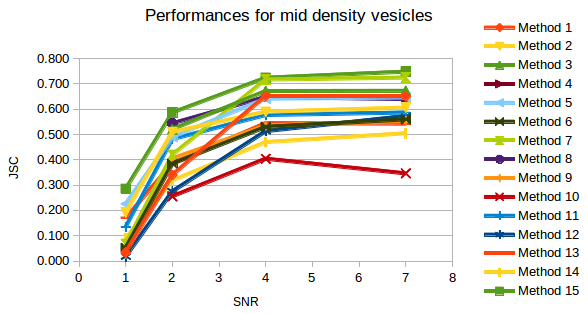}
\\
\includegraphics[width=8cm]{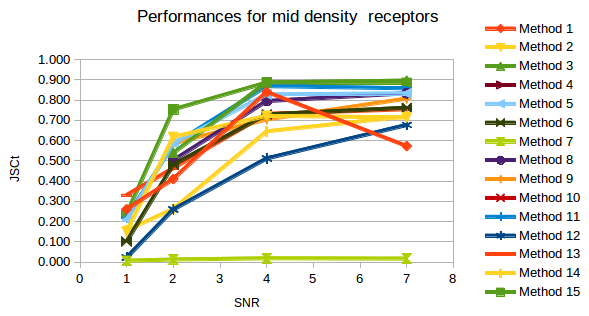}
&\includegraphics[width=8cm]{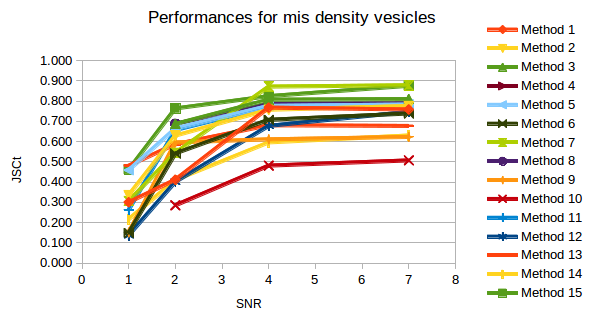}
\\
\includegraphics[width=8cm]{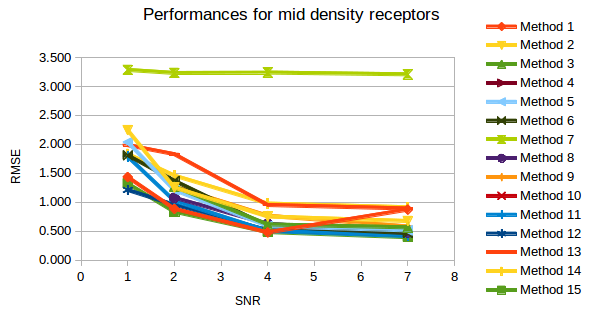}
&\includegraphics[width=8cm]{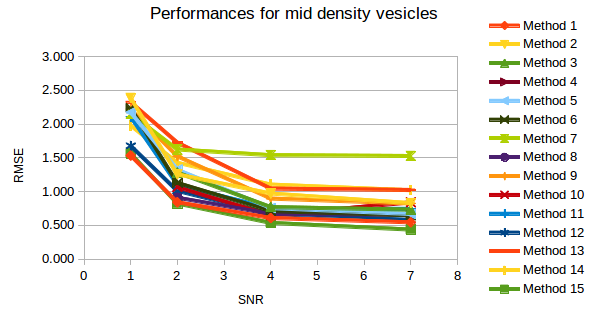}
\\ (a) & (b)
\end{tabular}
\caption{Performance measures for mid-density particle scenarios: (a) RECEPTORS, (b) VESICLES. From up to down are show the metrics $\alpha$, $\beta$, $JSC$, $JSC_t$ and $RMSE$. Method 15 is the proposed approach.}
\label{midDensity}
\end{figure}

\begin{figure}[H]
\begin{tabular}{ccrrrrrrrrrrrrrrr}
\includegraphics[width=8cm]{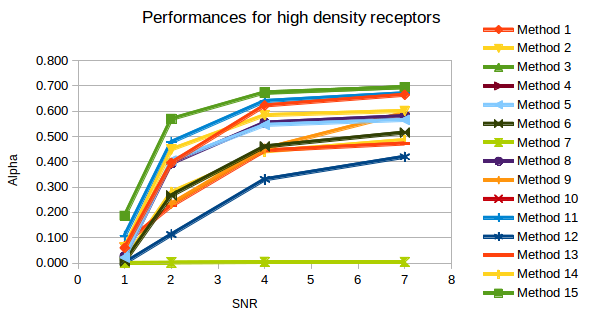}
&\includegraphics[width=8cm]{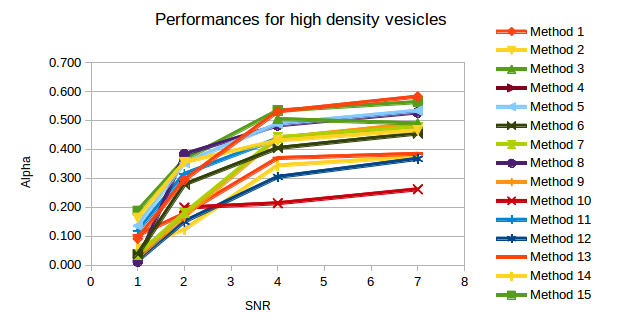}
\\
\includegraphics[width=8cm]{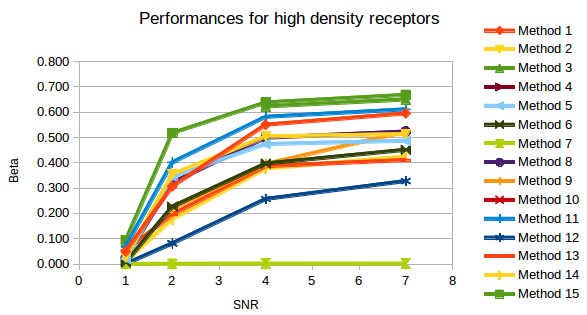}
&\includegraphics[width=8cm]{highDensityVesicleBeta.png}
\\
\includegraphics[width=8cm]{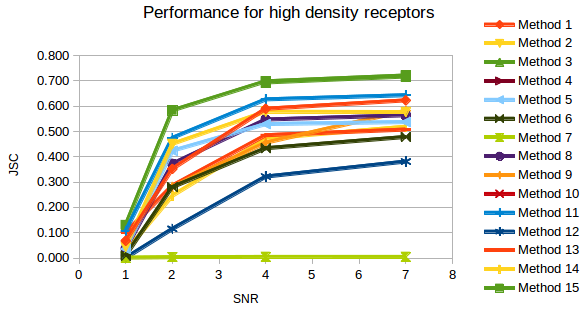}
&\includegraphics[width=8cm]{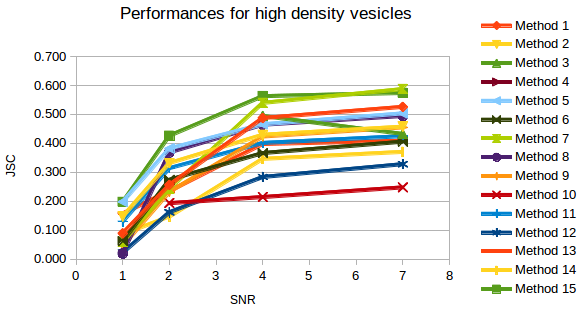}
\\
\includegraphics[width=8cm]{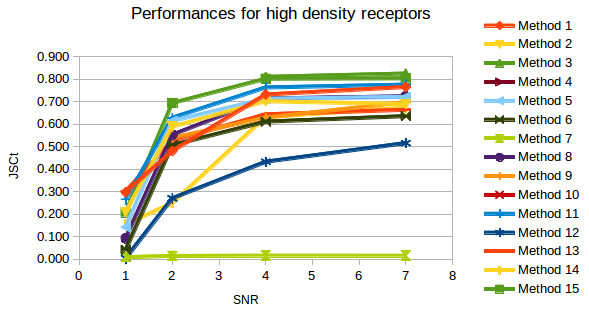}
&\includegraphics[width=8cm]{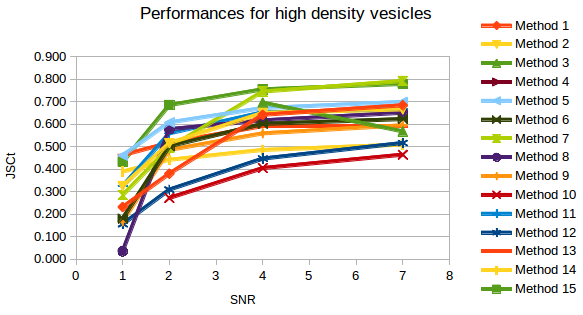}
\\
\includegraphics[width=8cm]{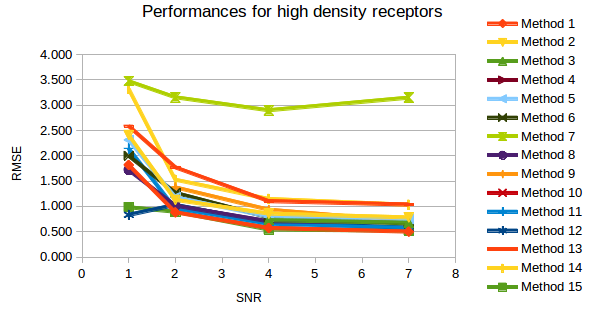}
&\includegraphics[width=8cm]{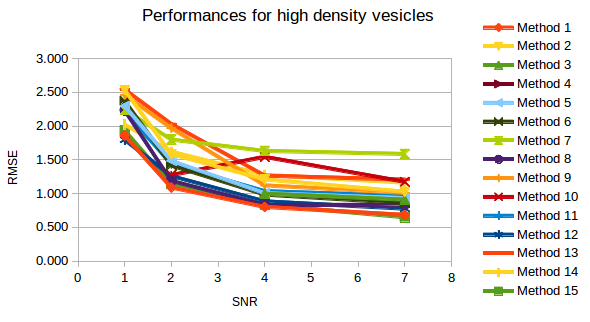}
\\ (a) & (b)
\end{tabular}
\caption{Performance measures for high-density particle scenarios: (a) RECEPTORS, (b) VESICLES. From up to down are show the metrics $\alpha$, $\beta$, $JSC$, $JSC_t$ and $RMSE$. Method 15 is the proposed approach.}
\label{highDensity}
\end{figure}

The methods for particle detection used in the challenge can be roughly grouped in four classes (see Table \ref{tab:14methods}): 1) thresholding or local maxima selection (methods 2,3,4,9); 2) linear (Gaussian, Laplacian of Gaussian and difference of Gaussian) and nonlinear model fitting including morphological processing (methods 6,7,11,12,13,14) 4) centroid estimation scheme (method 1) or a combination of them (methods 5,8,10). The methods for temporal linking include deterministic (method 1, 6, 8, 9, 12, 13,) and  probabilistic approaches (methods 2,3,4,5,7,10,11,14) and they were introduced in section \ref{sec:relatedWorkTRacking}.
In Fig.\ref{lowDensity}, Fig.\ref{midDensity}, and Fig.\ref{highDensity}  comparisons are shown for low, mid and high particle density scenarios respectively.  The left and the right columns show the performances in terms of the five measures defined above for RECEPTORS and VESCICLES respectively as a function of the SNR. As it can be observed, the performaces of all methods, within a given scenario, depends on particle density and SNR.
Analyzing trends, it emerges that generally the performances of the a-contrario approach, called Method 15 in the legend, decreases less strongly when the SNR goes from 4 to 2  with respect other methods. The metrics  $\alpha$ and $\beta$ differ slighly independently on the particle density and on the kind of particle motion (Brownian for VESICLE and Brownian switching to directed motion models for RECEPTORS), meaning that the a-contrario method gives a low number of spuriuos tracks. As general trend, these measures for the a-contrario approach decrease while increasing particle density, even if they are above the state of the art. This can be understood considering that the particle detection algorithm assumes that the minimal distance between two particles is at least of one pixel, an assumption that often does not hold in high density scenario. The Jaccardi similarity index for positions is clearly showing the best performances even for high particle density and low SNRs, for both VESICLE and RECEPTORS. This can mainly be due to the low number of spurious  tracks, which are penalized by this measure, indicating that the tracking algorithm is robust to The Jaccard similarity coefficient for entire tracks has in general higher value for RECEPTORS than for VESICLES with respect to other methods. This means that the a-contrario particle temporal linking is able to cope with the more complex dynamics of these particles better than competing methods. 
The localization accuracy, expressed in terms of RMSE is comparable with Method 12 of the state of the art which uses parabolic fitting for particle detection. However, it should be taken into account that this  measure is an average over all corrected matching pairs, which are not too much for this method, as demonstrated by its performances in terms of $\alpha$.  In general, the better performances of the a-contrario approach are mainly due to its control on the number of false alarms, which are penalized in 3 out of 5 the evaluation criteria.

\begin{center}
\captionof{table}{Detection and temporal linking methods compared in \cite{Chenouard2014Objective}  }
\begin{tabular}{|p{0.5cm}|p{3.5cm}|p{3.7cm} |p{1.8cm}| }
 \hline
 Met. & Detection approach  & Linking approach  & Refs.\\
 \hline
 1  & Iterative intensity-weighted centroid calculation &  Combinatorial tracking (deterministic) & \cite{sbalzarini2005feature}\\
 2  & Adaptative local-maxima selection & Multiple hypothesis tracking (probabilistic)  & \cite{coraluppi2011multi,coraluppi2004recursive}\\
 3  & Maxima after thresholding two-scale wavelet products &  Multiple hypothesis tracking (Probabilistic) & \cite{chenouard2009multiple,chenouard2009multiple2,olivo02}\\
 4  & Adaptive Otsu Thresholding & Multitemporal association tracking &  \cite{winter2012axonal,winter2011vertebrate}\\
 5  & Thresholding + centroid calculation & Kalman filtering + probabilistic data association & \cite{Godinez2009Deterministic,godinez2011tracking}\\
 6  & Lorentzian function fitting to structures above noise level  & Dynamic programming & \cite{rink2005rab} \\
 7  & Gaussian mixture model fitting & Multiple hypothesis tracking & \cite{liang2010tracking}\\
 8  & Watershed-based clump splitting and parabola fitting & Viterbi algorithm on state-space representation & \cite{magnusson2012batch,yin2012understanding}\\
 9  & Maxima with pixel precision  & Nearest neighbor + global optimization & \cite{husain2012software,casuso2012characterization}\\
 10 & Histogram-based thresholding and Gaussian fitting & Gaussian template matching & \cite{rousseeuw2005robust,hager2004multiple,schunck1989computing}\\
 11 & Gaussian fitting & Sequential multiframe assignment & \cite{olivo02,thompson02,shafique2005noniterative}\\
 12 & Parabolic fitting to localized maxima & Linear assignment problem & \cite{lowe2004distinctive,jaqaman2008robust} \\
 13 & Watershed-based clump splitting  & Nearest neighbor & \cite{crocker1996methods,celler2013single}\\
 14 & Morphological opening-based clump splitting & Nearest neighbor + Kalman filtering & \cite{ku2009morphological,ku2007automated}\\
 \hline
\end{tabular}
\label{tab:14methods}
\end{center}

\newpage

\newpage

\section{Conclusions}
\label{conclusions}
This paper has shown the advantages of the a-contrario framework for the spatial detection and tracking of near-circular particles in fluorescent time-lapse images.
Comparative evaluations under different biological scenarios and varying experimental conditions, both of the particle detection method and of particle detection followed by temporal linking, have shown that the proposed approach outperforms the state-of-the-art in terms well established performance measures. In addition to better performances for very low SNR, the a-contrario approach provides three additional advantages: a rigorous control of false detections in pure noise, which is important to avoid the corruption of quantitative analysis in biological data; low sensitivity to parameters changes; no need of a costly training stage that, could possibly introduce a bias.  These characterisitcs make the proposed algorithms particularly suited to be used by biologists. The ImageJ plugin of the particle detection algorithm can be found online at http://fluobactracker.inrialpes.fr/.
Future work will extend this framework to handle 3D data and elongated particles. 


\begin{acknowledgements}
This work was partially funded by the French National Research Agency (ANR) under contract ANR-09-PIRI-0030-03. The first author would like to thank two anonymous reviewers for their constructive comments that greatly contributed to improving the final version of the paper.
\end{acknowledgements}
\bibliographystyle{spbasic}  
\bibliography{biblio}                

\end{document}